\documentclass[journal]{IEEEtran}

\usepackage{graphicx}
\usepackage{subfigure}

\begin{document}

\title{Rethinking the Artificial Neural Networks: A Mesh of Subnets with a Central Mechanism for Storing and Predicting the Data}

\author{Usman Ahmad, Hong Song, Awais Bilal, Shahid Mahmood,  Asad Ullah, Uzair Saeed 

\thanks{Usman Ahmad and Hong Song are with the School of Computer Science and Technology, Beijing Institute of Technology, Beijing 100081, China (e-mail: usmanahmad@bit.edu.cn ; anniesun@bit.edu.cn).}
\thanks{Awais Bilal is with the School of Electrical Engineering and Computer Science, National University of Sciences and Technology, Islamabad 44000, Pakistan (e-mail: 13msccsabilal@seecs.edu.pk).}
\thanks{Shahid Mahmood is with the School of Computing, Electronics and Mathematics, Coventry University, Coventry, UK (e-mail: mahmo136@coventry.ac.uk).}
\thanks{Asad Ullah is with the School of Information and Electronics, Beijing Institute of Technology, Beijing 100081, China (e-mail: engr.asad@bit.edu.cn).}
\thanks{Uzair Saeed is with the School of Computer Science and Technology, Beijing Institute of Technology, Beijing 100081, China (e-mail: uzairsaeed@bit.edu).
}}

\markboth{SUBMITTED TO IEEE TRANSACTIONS ON NEURAL NETWORKS AND LEARNING SYSTEMS}%
{Ahmad \MakeLowercase{\textit{et al.}}: Rethinking the Artificial Neural Networks: A Mesh of Subnets with a Central Mechanism for Storing and Predicting the Data}

\maketitle

\begin{abstract}

The Artificial Neural Networks (ANNs) have been originally designed to function like a biological neural network, but does an ANN really work in the same way as a biological neural network? As we know, the human brain holds information in its memory cells, so if the ANNs use the same model as our brains, they should store datasets in a similar manner.  The most popular type of ANN architecture is based on a layered structure of neurons, whereas a human brain has trillions of complex interconnections of neurons continuously establishing new connections, updating existing ones, and removing the irrelevant connections across different parts of the brain. In this paper, we propose a novel approach to building ANNs which are truly inspired by the biological network containing a mesh of subnets controlled by a central mechanism. A subnet is a network of neurons that hold the dataset values. We attempt to address the following fundamental questions: (1) What is the architecture of the ANN model? Whether the layered architecture is the most appropriate choice? (2) Whether a neuron is a process or a memory cell? (3) What is the best way of interconnecting neurons and what weight-assignment mechanism should be used? (4) How to incorporate prior knowledge, bias, and generalizations for features extraction and prediction? Our proposed ANN architecture leverages the accuracy on textual data and our experimental findings confirm the effectiveness of our model. We also collaborate with the construction of the ANN model for storing and processing the images.

\end{abstract}


\IEEEpeerreviewmaketitle

\section{Introduction}
The biological neural network is perhaps the most complex and advanced biological system on the planet and we know very little about it as an information processing organ. A number of questions remain unanswered about how the information is encoded and transferred from neuron to neuron or from network to network of the neurons. Artificial Neural Network (ANN) is one of the most influential innovations in the domain of computer vision, inspired by the biological neural network systems. ANNs enable computers to learn from examples that are available to them in the form of data. ANNs have turned out to be a key to cramming knowledge into computers, which ultimately enables intelligence. Feeding knowledge into computers has been a key challenge for ANN approaches, as human beings have a natural ability to learn and acquire new skills on their own. With the help of ANNs, we are able to program computers to acquire new knowledge in a very similar way as human beings do. 

The principles underpinning the flight of birds led to the discovery of wonderful laws of aerodynamics, which ultimately led to the invention of aircrafts. Similarly, the principles underpinning the human intelligence can potentially be taught to computers by means of ANNs, for example. However, unlike observing physical properties of some phenomenon (such as how a bird is able to fly), understanding human intelligence may not be as straightforward, as it does not have any physical characteristics to be perceived. Therefore, we have to turn to the natural yet powerful tool of learning: \textit{observation} combined with the scientific investigation which will help us to build our ANN model.

Human organs, such as ears, eyes, nose, tongue, and skin receive input from the outside world, encode into signals and relay these signals to the human neurons to be processed. The signal processed through millions of subsequent neurons to arrive at a conclusion. We examine the functioning of the human neural network by observing the organs that perceive input from the outside world and the results concluded by the human neural network, similar to the behavioral or black-box testing of a system. As far as the internal structure of the biological neural network is concerned, we relate our observation results on the human organs to the scientific investigation and connect the dots. 


In this paper, we propose a simple, yet powerful ANN architecture inspired by the biological neural network. Our proposed ANN architecture contains a mesh of subnets, where a subnet is a group of neurons. Memory management is one of the main functions of the human brain, similarly the neurons in our ANN model act as memory cells to hold the dataset values. Usually, we arrange our text datasets in the form of a table; in contrast, in our ANN model, the dataset is organized in the form of subnets wherein a value of a specific data type is stored in a separate subnet. Neurons in the ANN are distributed by the subnets, and a group of neurons in a subnet contains a particular type of data.

Our propose ANN model is under the control of a central mechanism, same like the heart in human body. This central mechanism is responsible for making decisions, controlling and monitoring the entire workings of the ANN. Recent studies have revealed that the brain and heart are closely connected and work together for processing the signals, and that the heart also contains neurons. 
Recent work in the new field of neurocardiology has established that the heart is a sensory organ and an information encoding and processing center \cite{mccraty2009coherent}. The HeartMath Institute Research Centre has been working for 27 year and explored how the activity of the heart influences our perceptions, emotions, intuition and health. In \cite{mccraty2001science}, the authors conclude that the heart is the source of emotions, courage, and wisdom, and it has the ability to influence our perceptions, feelings, intuition and health. According to Dr. David Paterson at Oxford University, the heart also contains neurons similar to those in the brain. The heart and the brain are connected and brain is not the only organ that produces emotions \cite{larsen2016sympathetic}.

To the best of our knowledge, this is the first work that provides an ANN model under the control of a central mechanism based on a combination of behavioral observation of human intelligence and scientific research on human neurons. The rest of the paper is organized as follows. We illustrate the key components of our proposed ANN model in Section II. In section III, we present the ANN's architecture for text data processing. In section IV, we evaluate our model through Iris dataset example. In section V, we present the foundation of our proposed ANN's for image processing. Section VI, reviews the literature review. We answer the important questions about our proposed model in Section VII. Finally, we present conclusions and directions for future in Section VIII.


\section{Proposed Architecture for the ANN}

In this section, we present the basic components of our proposed ANN model followed by some examples to explain the training and testing phases. In \cite{simon2015organic}, the authors have claimed to build artificial neuron containing no living parts, but it is still capable of mimicking the functions of human neuron cells and communicating in the same way as our biological neurons do. Although we are not able to mimic the exact working of human neurons using bioelectronics, we can, nevertheless, implement our proposed ANN using the binary method to store the data in the neurons. 

\begin{figure}[!t]
    \centering

        \includegraphics[width = 3.5in]{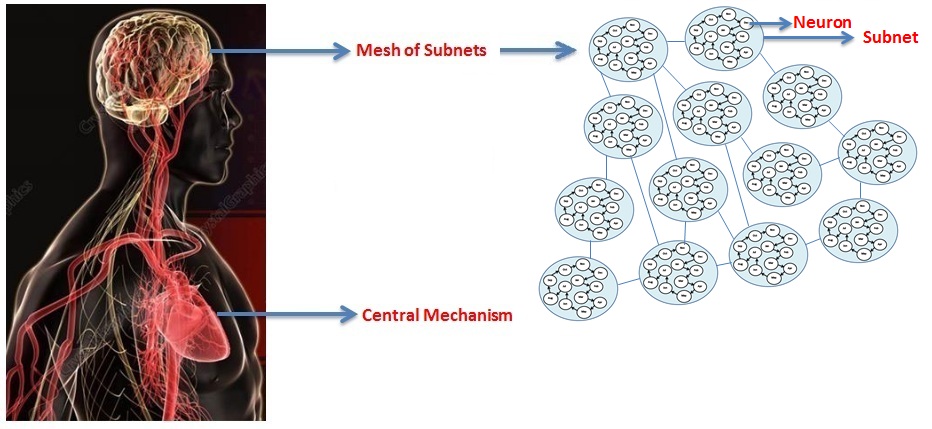}
     
     \caption{Human biological system and our proposed ANN model}
        \label{Fo}
\end{figure}

\subsection{Mesh of Subnets}
Neurons in an ANN must not be in a layered format, instead, they should be interconnected in a way that we would like to refer as the \textit{mesh of subnets}. In our proposed architecture, ANN is a combination of interconnected subnets, where a subnet is a collection of neurons. A subnet is basically the distribution of neurons in an ANN. A neuron may belong to one or more subnets and more likely to be interconnected with some or all neurons of its subnet(s). A neuron may also be interconnected to some other subnet's neurons. In our ANN, the dataset is stored in the form of subnets and a particular type of data is stored in its own subnet. An abstract view of our ANN model is depicted in Figure \ref{Fo}.

\subsection{Neuron}

In our proposed ANN, a neuron is a memory cell and holds the data that is supplied to the ANN for both training and testing purposes. In some cases, a neuron can contain a function in its memory; it is further discussed in section III. Neurons are interconnected and each connection has a specific weight value. New connections are established between the neurons, become stronger or weaker, or removed along with the connected neurons based on the frequency of input data flows between neurons during the training and testing processes. This is similar to the process of how a human brain learns new things and adds experiences to those learnings to draw the conclusions on the incidents occurring in our lives. 

When a connection is created between two neurons of different subnets, then most of the time, first a connection is established between the subnets then between the neurons of the subnets. The weight value defines how strong binding two neurons have with each other as well as the distance of a neuron from other neurons. In other words, the weight defines the location of a neuron in an ANN. A neuron can move from one subnet to another existing or newly formed subnet.

\subsection{Subnet}
A subnet is a distribution of neurons in an ANN, whereby each neuron is connected to the central mechanism. New subnets along with the neurons are created during the training and testing phases. The subnets are interconnected, and each connection has a weight value. The connection and the corresponding weight between two subnets depend upon the frequency of input data flow between the neurons of the subnets during training and testing process. A new subnet can be created by merging two or more subnets. Conversely, an existing subnet can be split into more than one subnets. A subnet may have child and parent subnets. A subnet can have more than one related subnets, called super subnet. As every person may have their own way to remember and recognize things; in a similar way, every ANN may have different subnets and neuron distributions. In the case of textual data, the subnet creation can be formalized. One way to achieve this may be the number of subnets equal to the number of attributes in the dataset. So, we can also declare the number of subnets before training. If there is a human intervention involved in the creation and specification of subnets, we can refer to it as a \textit{semi-trained} ANN model.

\subsection{A Central Mechanism}

An ANN must have the main decision making, central and external mechanism to control and monitor the functioning of the ANN. The central mechanism takes input and decides how deeply (i.e. how many subnets containing datasets, prior knowledge and biases will be included in the processing) to process the test data in the ANN. Each neuron in the ANN is connected to the central mechanism through a subnet, and occasionally directly. This ensures that each neuron is in connection and has direct access to all the neurons in the ANN through the central mechanism. The central mechanism monitors the input data being processed and move from one neuron/subnet to the next.  The central mechanism also provides a direct path for input data to move from one neuron to another in the entire ANN. 

Additionally, the central mechanism also contains neurons and subnets in the same way as the ANN for temporarily holding input and output data as well as the information of the ANN's subnets for communicating with the ANN memory to forward the input to appropriate subnets and obtain results from the ANN. Creating new neurons, subnets, and establishing connections among them is yet another responsibility of the central mechanism. As can be noticed, the central mechanism has multiple crucial roles to play in our proposed ANN model, we further elaborate the working of it by means of some examples.

\subsection{Connections and Weights Assignment}

As described earlier, the central mechanism is responsible for creating new neurons and subnets and establishing connections among them. During the training and testing phases, the central mechanism splits the input data into single storable values to put into the neurons and defines the data type of each value in order to determine the subnets for holding the data values. The mechanism creates a new neuron in the respective subnet stores the value into it. If the value already exists in that specific subnet, it only updates the weight value and does not repeat the value in order to avoid duplicate values. 
No two neurons in a subnet can contain the same value. 

If a subnet does not already exist, the central mechanism creates one followed by the new neuron(s) to store the values. Based on the flow of incoming data, the mechanism creates connections between neurons/subnets and assigns an initial weight value for each connection. The direction (depicted by an arrowhead) of the connection specifies the sequence of input data values. Once weight has been assigned, it is updated on each occurrence of the same input data values making the connection stronger on each iteration. It is important to note that the direction of the connection between neurons shows the sequence of input, it does not restrict the flow of data in another direction. This is also valid in the testing and feedback loops. Our proposed ANN model contains the following types of connections:

\begin{enumerate}
\item Neuron-to-neuron 
\begin{enumerate}

\item between the neurons of the same subnet in an ANN
\item between the neurons of the same subnet in central mechanism
\item between the neurons of the different subnets in ANN
\item between the neurons of the different subnets in central mechanism
\item between a ANN's subnet neuron and central mechanism's subnet neuron

\end{enumerate}

\item Subnet-to-subnet

\begin{enumerate}

\item between the subnets in ANN
\item between the subnets in central mechanism
\item between the ANN's subnet and central mechanism's subnet

\end{enumerate}

\item Between the neurons and subnets

\begin{enumerate}

\item between the neuron of ANN's subnet and other subnet in ANN
\item between the neuron of central mechanism's subnet and other subnet in central mechanism
\item between the neuron of ANN's subnet and other subnet in central mechanism
\item between the neuron of central mechanism's subnet and other subnet in ANN

\end{enumerate}
\end{enumerate}

Each connection in our proposed model belongs to one of the types listed above and has a particular label, if however there is no label then it can be set to null. Moreover, a connection may have more than one labels. The label defines the nature of the relationship between the neurons. The connections can be bidirectional if the input data flow occurs in both directions. Practical examples of connection labels are provided in the following section. 

\section{Proposed ANN's Approach to Textual Data Processing from Scratch}

In this section, we explain the training and testing process of our proposed ANN model using some example. For a better understanding and simplicity, we select a small dataset containing only 30 records (see Appendix A) for depicting the ANN structure by using figures. This is followed by another example that includes the use of Iris dataset. The latter is presented in the next section. Our ANN model can accommodate larger datasets equally well by expanding itself according to the size of the dataset. Our sample dataset consists of a log of insulin doses taken by a diabetic patient over a period of two weeks. The dataset contains three attributes: date, time and the volume of insulin dose. We build an ANN model that predicts the amount of insulin to be injected on a given date and time.

\subsection{Prior Knowledge and Feature Extraction}

One record is inserted at a time and all the thirty records are supplied to the ANN model for training. The central mechanism takes this input data and creates subnets, neurons, and connections among them. Before training the model, we must have prior knowledge of the dataset.
The date attribute of our dataset is made up of a day and a month. Similarly, the time attribute consists of hours and minutes. The third and final attribute volume of insulin is represented by a number. In this particular case, we need prior knowledge of days, months, hours, minutes, and numbers so we have three subnets for holding values of the prior knowledge. Subnets 1, 2, and 3 are given below:

\begin{itemize}
\item Subnet 1: Contains positive integer number system i.e. 1, 2, 3... 
\item Subnet 2: Contains the names of months
\item Subnet 3: Contains the time structure, i.e. 24 hours and 60 minutes 
\end{itemize}

\begin{figure}[!t]
    \centering
    \subfigure[Subnet 1: numbers]
    {
        \includegraphics[width = 2in]{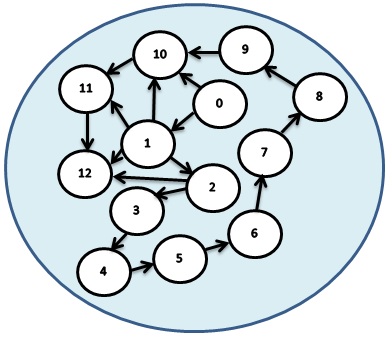}
        \label{f1_1}
    }
    \\
    \subfigure[Subnet 2: names of months]
    {
        \includegraphics[width = 2in]{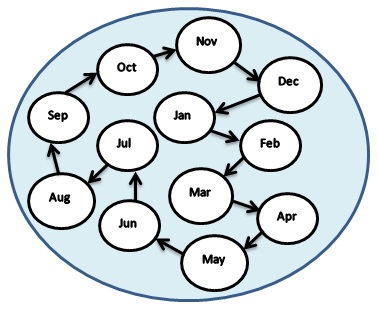}
        \label{f1_2}
    }
    \\
        \subfigure[Subnet 3: time structure]
        {
            \includegraphics[width = 1.3in]{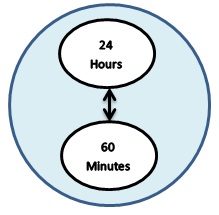}
            \label{f1_3}
        }
     \caption{Subnet 1, 2, and 3 containing the prior knowledge}
        \label{f1}
\end{figure}

Figure \ref{f1_1} illustrates positive integer number system. The connections between the neurons are in accordance with the sequence of occurrence of the input values and correspondence between the input values. ANN must have some limited prior knowledge for practical implementation of ANNs. The values in the subnet 1 starts from 0 to the maximum value exist in our dataset, which is 36. Figure \ref{f1_1} depicts the architecture for 0 to 12 values due to space restrictions and it will grow in the same way up to the value 36. Subnet 1 has 12 neurons to hold values from 0 to 12: each value per neuron. We will insert the values 0, 1, 2, 3, 4, 5, 6, 7, 8, 9, 10, 11, 12 to subnet 1 is a sequence, similar to a child who learns the counting in the school. First neuron containing value 0 connected to the second neuron containing value 1 which is further connected to third neurons containing value 2 and so on. The neuron contains value 9 is connected to the neuron contains value 10. Neuron 10 is a combination of two values i.e. 1 and 0, so also connected with two neurons having values 1 and 0. The arrow from neuron 9 to 10 indicates the sequence which means that 10 is after 9. The label of the connection between them is \textit{less than}, which denotes that 9 is less than 10. The label of connection from neuron 1 to 10 is \textit{part 1 of 2} and the label of connection from neuron 0 to 10 is \textit{part 2 of 2}, which denotes the neuron 10 is a combination of two neurons that are 1 and 0.

\begin{figure}[!t]
    \centering

        \includegraphics[width = 3in]{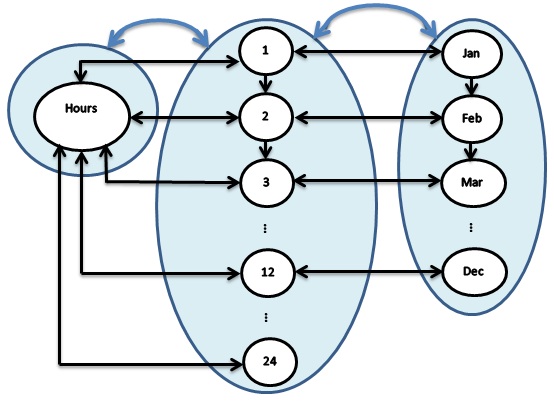}
     
     \caption{Prior knowledge subnets 1, 2, and 3 interconnections}
        \label{f2}
\end{figure}

Subnet 2 in Figure \ref{f1_2} contains the names of months, from January to December. One neuron containing value June is enough for our dataset because all values in the date attribute belong to the month of June. Subnet 3 in Figure \ref{f1_3} contains the time structure, i.e. hours and minutes. We did not include the seconds as the format of time attribute in our dataset does not have seconds. Prior knowledge subnets 1, 2, and 3 are interconnected as shown in Figure \ref{f2}. Neurons 1 to 12 in subnet 1 are connected to the neurons Jan to Dec in subnet 2, respectively. Similarly, the human brain holds the knowledge of 12 months in a year. We have 24 hours in a day so the neuron hour in subnet 3 is connected to neurons 1 to 24 in subnet 1. Similarly, the neuron minutes will be connected to 60 values.

\begin{figure}[!t]
    \centering
   
        \includegraphics[width = 2.5in]{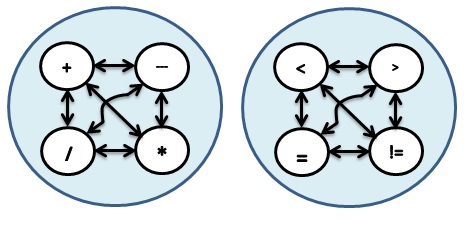}

     \caption{Subnet 4 and 5 containing the operations for generalization to extract the features}
        \label{f3}
\end{figure}

As we mentioned earlier that a neuron can also contain the functions. We need to perform some operations to extract the features. We have the subnets 4 and 5 containing relevant functions for generalization to extract the features and patterns. Subnet 4 and 5 contains the arithmetic operators and relational operators, respectively, drawn in Figure \ref{f3}. The connections between the neurons of subnets 4 and 5 are bidirectional because the data flows in both directions during the training and testing phases. In the training phase, these operations are selected from neurons and are assigned as labels to the connections between neurons/subnets; whereas, in the testing phase, these operations are used to compare the test record values with the neuron values. Subnets 4 and 5 as outlined below:

\begin{itemize}
\item Subnet 4: Contains arithmetic operators 
\item Subnet 5: Contains relational operators 
\end{itemize}

The subnets 1, 2, and 3 holding prior knowledge are interconnected with the subnets 4 and 5 containing the operators. 

\subsection{Training}

We have three attributes in our dataset, and correspondence between them. So, we have the subnet 6, 7, and 8 to hold the input data as given below:

\begin{itemize}
\item Subnet 6: Contain the date that is first attribute of dataset
\item Subnet 7: Contain the timestamps that is second attribute of dataset
\item Subnet 8 (target subnet): Contain the value of insulin amount 
\end{itemize}

\begin{figure}[!t]
    \centering
    \subfigure[Subnet 6: date attribute of dateset]
    {
        \includegraphics[width = 2in]{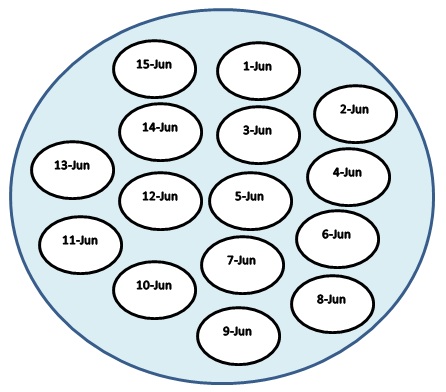}
        \label{f4_1}
    }
    \\
    \subfigure[Subnet 7: time attribute of dateset]
    {
        \includegraphics[width = 2in]{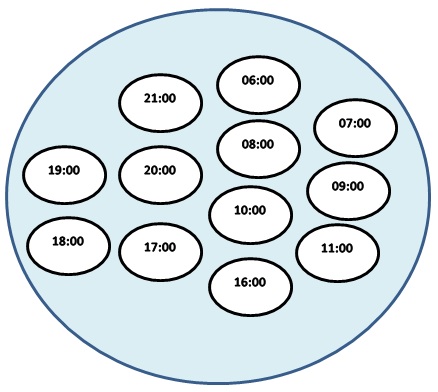}
        \label{f4_2}
    }
    \\
        \subfigure[Subnet 8: inulin amount attribute of dateset]
        {
            \includegraphics[width = 1.5in]{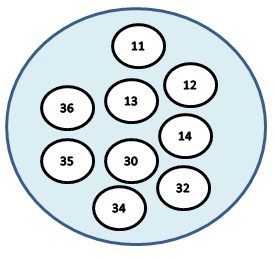}
            \label{f4_3}
        }
     \caption{Subnet 6, 7, and 8 holding the training dataset}
        \label{f4}
\end{figure}

In our dataset, we have 15 unique values in the date attribute, 12 unique timestamps in the time attribute and 9 unique insulin dose values in the third attribute, so subnets 6, 7, and 8 have 15, 12 and 9 neurons, respectively. Figure \ref{f4} illustrates the subnet architecture of subnets 6, 7, and 8. Since the subnet 8 contains the data to be predicted, i.e. insulin dose value, so the subnet 8 is called target subnet.

\begin{figure}[!t]
    \centering
    \subfigure[Subnet 6 (center) connections with prior knowledge subnets 1 and 2]
    {
        \includegraphics[width = 3.25in]{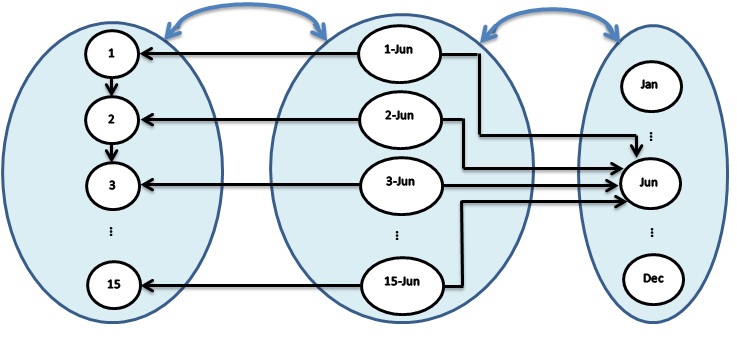}
        \label{f5_1}
    }
    \\
    \subfigure[Subnet 7 (center) connections with prior knowledge subnets 1 and 3]
    {
        \includegraphics[width = 3.25in]{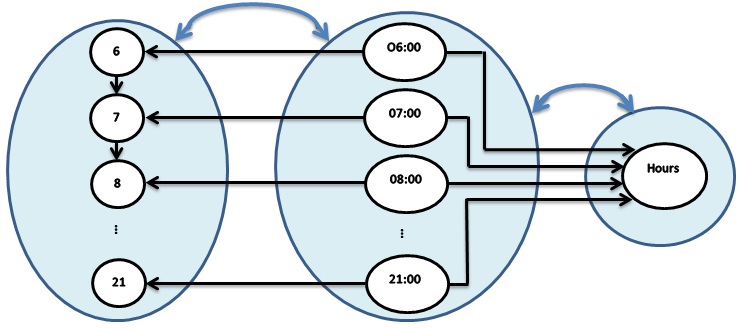}
        \label{f5_2}
    }
    \\
        \subfigure[Subnet 8 (center) connections with prior knowledge subnet 1]
        {
            \includegraphics[width = 2.5in]{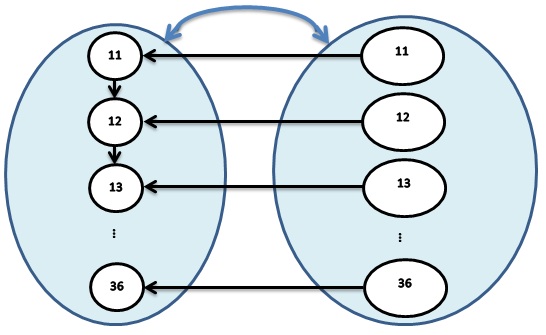}
            \label{f5_3}
        }
     \caption{Connection between dataset attributes subnet's neurons and prior knowledge subnet's neurons}
        \label{f5}
\end{figure}

Figure \ref{f5} depicts the connections between subnets 6, 7, and 8 containing dataset attributes and subnets 1, 2, and 3 holding prior knowledge. Subnet 6 contains the date attribute and has connections with prior knowledge subnets 1 and 2, as shown in Figure \ref{f5_1}. Neurons 1-Jun to 15-Jun of subnet 6 are connected to neurons 1 to 15 of subnet 1, respectively. All neurons of subnet 6 are connected to neuron Jun of subnet 2. Subnet 7 contains time attribute of the dataset and have connections with prior knowledge subnets 1 and 3, as shown in Figure \ref{f5_2}.  All neurons 06:00 to 21:00 of subnet 7 are connected to the respective neurons of subnet 1. All neurons of subnet 7 are also connected to neuron hours of subnet 3. Similarly, the neurons of target subnet 8 contains insulin dose value connected with prior knowledge subnet 1, as shown in Figure \ref{f5_3}. Prior knowledge subnets 1, 2, and 3 are also connected to each other, as mentioned above and shown in Figure \ref{f2}.


\begin{figure}[!t]
    \centering

        \includegraphics[width = 3.25in]{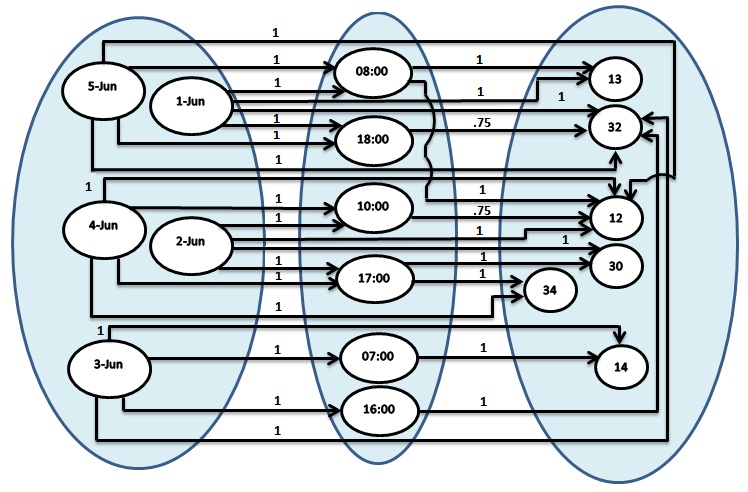}
     
     \caption{Subnet 6, 7, and 8 (left to right) structure for the first 10 records in our insulin log dataset}
        \label{f6}
\end{figure}

Figure \ref{f6} depicts the subnets holding the training data of the first 10 records in our dataset. In the first 10 records, we have 5 unique values in the first attribute and 6 unique values in both second and third attributes, so subnet 6 has 5 and subnet 7 and 8 have 6 neurons. For this example, we have assigned the initial weight value 1 to each connection between the neurons and decreased by .25\% on each iteration. The label of all connections between these neurons is “If...Then”. In the first 10 records of our dataset, the values 18:00 and 10:00 are repeated in time attribute and the values 32 and 12 repeated in insulin dose attribute. So, the weight value of connection between these neurons is 0.75. Although we did not mention the weights and labels of connections for subnets 1, 2, 3, 4, and 5, but they do have the same mechanism of weight assignment and label like subnets 6, 7, and 8 do, as shown in \ref{f6}.

\subsection{Testing}
We trained our model with the first 10 records and picked the 11th record as a test record, so we have to predict the insulin amount against the 6-Jun and 11:00. The central mechanism is responsible to take input and forward it to the appropriate subnets. The neurons in the central mechanism have information about the structure of ANN subnets. The central mechanism takes initiate from the base subnets containing the prior knowledge in order to find and approach the appropriate subnets to forward the input value. This is because the prior knowledge subnets are interconnected and further connected with the subnets containing dataset values. The date attribute value, i.e. 6-Jun, is placed in the subnet 6, and the time (i.e. 11:00) is entered in the subnet 7. Forwarding the input flow to the appropriate subnets is very important, which is monitored by the central mechanism. After putting the values into subnets, compare the input values with the neuron's values in the subnets and generate a result (yes/no). If the value does not match, then find and select the nearest value neuron. The first value (6-Jun) does not exist in subnet 6, so we picked 5-Jun which is the nearest value.

Next step is to find all connected neurons with this selected neuron (5-Jun) meeting two conditions: \textit{first, it must belong to a subnet (except the target subnet) which contains dataset attribute values, second, it must be connected to the same neurons in the target subnet as our input (or nearest selected) value is connected.} So, our first selected neuron (5-Jun) has two connections meeting these two conditions; those are 08:00 and 18:00 neurons in subnet 7. The neurons 08:00 and 18:00 belong to the dataset subnet 7 and connected to the same neurons as the neuron 5-Jun is connected; those are 12 and 32, respectively. Now pick the second input value (11:00) and compare with the selected connected neurons from subnet 7 (08:00 and 18:00). If the value does not match, then find and select the nearest value neuron. The second input value i.e. 11:00 compared with selected neurons that are 08:00 and 18:00 and the nearest value is 08:00. The 08:00 neuron is connected to the neuron 12 in target subnet, which is selected value against the first input value.

We repeat the process for each input values. The second value (i.e. 11:00) does not exist in subnet 7 so we picked 10:00 which is the nearest value. The neuron 10:00 has two connections meeting the conditions; those are 2-Jun and 4-Jun neurons in subnet 6. The neurons 2-Jun and 4-Jun are connected to the same neuron as the neuron 10:00 is connected; that is 12. The first input value (i.e. 6-Jun) is compared with selected neurons that are 2-Jun and 4-Jun and the nearest value is 4-Jun. The 4-Jun neuron is connected to the neuron 12 in target subnet; the neuron 12 is the selected value against the second input value. The value 12 is the selected against both, first (in the above paragraph) and second input value, so 12 in the predicted value of insulin amount against the values 5-Jun and 11:00.

Let us take the 12th record as a test record, we have to predict the inulin amount against 6-Jun and 17:00. The value 6-Jun is input to the subnet 6 and 11:00 is entered in the subnet 7. The first value (i.e. 6-Jun) does not exist in subnet 6, so we picked 5-Jun which is the nearest value and it has two connections fulfilling the criteria; those are 08:00 and 18:00 neurons in subnet 7. The neurons 08:00 and 18:00 are connected to the same neurons as the neuron 5-Jun is connected; those are 12 and 32, respectively. The second input value (i.e. 17:00) compared with selected neurons that are 08:00 and 18:00, and the nearest value is 18:00. The 18:00 neuron is connected to the neuron 32 in target subnet, which is selected value against the first input value. Repeating the process for the second input value, the second value 17:00 exists in subnet 7. The neuron 17:00 has two connections meeting the conditions; those are 2-Jun and 4-Jun neurons in subnet 6. The neurons 2-Jun and 4-Jun are connected to the same neurons as the neuron 17:00 is connected; those are 30 and 34, respectively. The first input value i.e. 6-Jun is compared with selected neurons that are 2-Jun and 4-Jun and the nearest value is 4-Jun. The 4-Jun neuron is connected to the neuron 34 in target subnet, which is selected value against the second input value.  The values selected against our first and second input are 32 and 34, respectively, and the average value is 33, that is the predicted value of insulin amount against the values 5-Jun and 17:00.

If a neuron belongs to more than one neuron, then we will consider the neuron having the minimum weight value. Bias is also stored in the neurons in a dedicated area of ANN. It is stored in the same way as the training data and used to process the test data. Neurons are interlinked with the neurons containing biased data, so they have an impact on the output. For example, if a person has taken an extra quantity of sugar then he should add 2 points of insulin amount in the routine insulin amount value. So, during the testing, the ANN model will add value 2 in the predicted value of insulin amount. If the testing results are true, then the test data value will be stored, and weight value will be updated in the same way as it was updated in the training phase.


%

%
  %

\section{Iris Dataset Example}	
In this section, we explain the training and testing process for our proposed ANN model through Iris dataset example.  The Iris flower dataset is introduced by the British statistician and biologist Ronald Fisher in 1936 \cite{fisher1936use}. The dataset consists of 50 samples from each of three species of Iris: setosa, virginica and versicolor. Four features were measured from each sample: the length and the width of the sepals and petals. Iris dataset attributes; sepal length, sepal width, petal length, and petal width contain decimal values and target attribute contains text data. So, we must have a subnet to store the decimal data as prior knowledge in the manner illustrated in Figure\ref{f1_1}, as the positive integer number system for insulin pump example. We have subnet 2 and 3 having arithmetic operators and relational operators for generalization to extract the features and patterns in the same manner as in subnet 4 and 5 of the insulin pump example, shown in Figure \ref{f3}. Subnet 1, 2 and 3 for Iris dataset are listed below:

\begin{itemize}
\item Subnet 1: Contains decimal values system i.e. 0.1, 0.2, 0.3...
\item Subnet 2: Contains arithmetic operators 
\item Subnet 3: Contains relational operators 
\end{itemize}

Iris dataset has five attributes and correspondence between them. So, we have the subnets 4, 5, 6, 7, and 8 to hold the dataset values. As we mentioned earlier that a subnet is a network of unique value neurons and duplicate values are not permitted. Iris dataset has 35, 23, 43, 22, and 3 unique values in sepal length, sepal width, petal length, petal width, and species, respectively.  So, subnets 4, 5, 6, 7, and 8 contain 35, 23, 43, 22, and 3 neurons, respectively.

\begin{itemize}
\item Subnet 4: sepal length
\item Subnet 5: sepal width
\item Subnet 6: petal length
\item Subnet 7: petal width
\item Subnet 8 (target subnet): species
\end{itemize}

The dataset attribute subnets have similar connections with prior knowledge subnet 1 as depicted in Figure \ref{f5_3} for the insulin pump example.

\subsection{Testing}

\begin{figure}[!t]
    \centering

        \includegraphics[width = 3.5in]{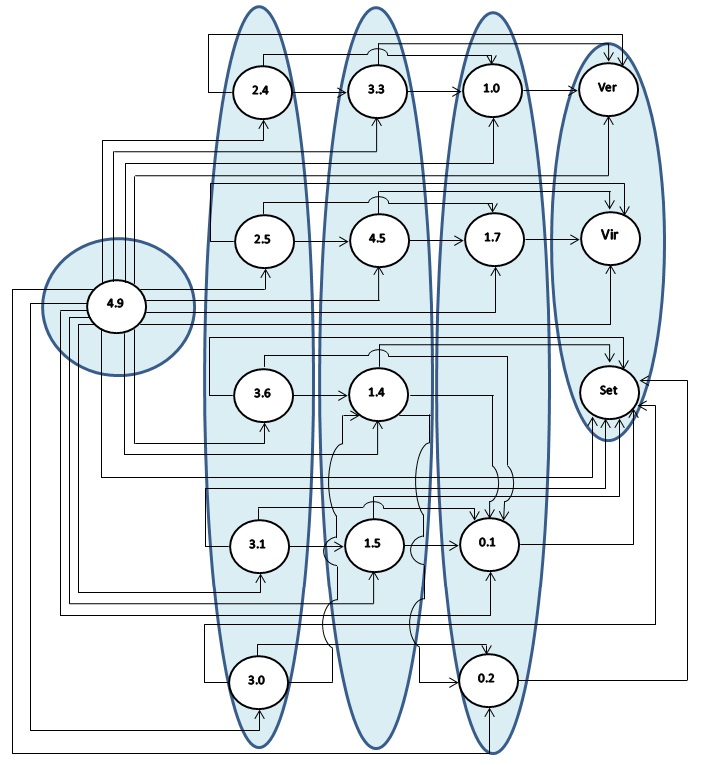}
     
     \caption{A snapshot of ANN architecture of subnet 4, 5, 6, 7, and 8 (from left to right) for 35th test record in Iris dataset}
        \label{f7}
\end{figure}

We randomly picked 35th record form Iris dataset for testing and rest of 149 records for training. The values in the 35th record are 4.9, 3.1, 1.5, 0.2, and setosa, respectively. A view of the ANN architecture of subnets 4, 5, 6, 7, and 8 for 35th test record in Iris dataset is illustrated in Figure \ref{f7}. The value 4.9 occurred 5 times in sepal length attribute (excluding the test record), so it has five corresponding values in each of the remaining four attributes. The attributes sepal width, petal length, petal width, and species have 5, 4, 4, and 3 unique values, respectively, against the value 4.9 in sepal length. The values 1.4 and 0.1 repeated in petal length, petal width, respectively. So, the subnet 5, 6, 7, and 8 containing data of sepal width, petal length, petal width, and species have 5, 4, 4, and 3 neurons respectively.

Central mechanism forwards the input values to the respective subnets. The sepal length attribute value 4.9 put in the subnet 4, sepal width value 3.1 to subnet 5, petal length value 1.5 to subnet 6 and petal width 0.2 to subnet 7. First input value 4.9 exists in subnet 4 and has total 13 connections: 5, 4, and 4 with subnets 5, 6, and 7 neurons, respectively, which meet two conditions (given in the insulin pump example). Now pick the second, third and fourth input values and compare them with the selected subnets 5, 6, and 7 neurons, respectively.

The second input value 3.1 compared with selected neurons in subnet 5 that are 2.4, 2.5, 3.0, 3.1, and 3.6 and value 3.1 is selected. The neuron 3.1 is connected to the neuron setosa in target subnet. The third input value 1.5 is compared with selected neurons in subnet 6 that are 1.4, 1.5, 3.3, and 4.5 and value 3.1 is selected. The neuron 1.5 is also connected to the neuron setosa in target subnet. Similarly, the fourth input value 0.2 is compared with selected neurons in subnet 7 that are 0.1, 0.2, 1.0, and 1.7 and value 0.2 get selected. The neuron 0.2  is also connected to the neuron setosa in target subnet. So, the setosa is the result of all the three values, i.e. the second, third, and fourth input values against the first input value. If more than one value results, then consider the neuron having the minimum weight value or having the maximum occurrence rate.

Repeat the process for each input values. The second value 3.1 occurred 10 times in sepal width attribute, so it has 10 corresponding values in each of the four attributes that are total 23 values. Now pick the first, third and fourth input values and compare them with the selected connected neurons against second input value in respective subnets. The first input value 4.9 is compared with 6 selected neurons in subnet 4 and the result is setosa in target subnet. The third input value 1.5 is compared with 9 selected neurons in subnet 6 and the result is setosa in target subnet. The fourth input value 0.2 is compared with 8 selected neurons in subnet 7 and the result is setosa in target subnet. So, the setosa is the result of all the three values, i.e. the first, third and fourth input values against the second input value.

The third input value 1.5 has occurred 12 times in petal length attribute, so it has 12 corresponding values in each of the four attributes. Similarly, the fourth input value 0.2 is occurred 28 times in petal width attribute, so it has 12 corresponding values in each of the four attributes.  Setosa is the result third and fourth input values. So finally, we obtain the value setosa against all four input values, which is an accurate value. We take the 94th record as a test case; it also gives the accurate result that is versicolor.


\section{Proposed ANN's Construction for Image Processing}

In image processing, an input image is analyzed, and as the output may either be an image, or a set of characteristics or parameters related to the image, which best describes the input image. This section explains how we store images in our ANN model, so they can be optimally used to carry out a comparison with a sample image and be useful in image recognition. 

As we mentioned earlier, we observe the nature's creation to design our ANN model and we particularly analyze the human eyes along with human neural network for image processing, as eyes are located close to the brain and eyeballs are directly connected to the brain through the optic nerve. A human eyeball has a black circle in the center with the white background. This contrast between black and white is very powerful and the base of how the human processes an image. We can also say black means "something" and white means "nothing". The human brain breaks the image into the smallest visible marks called X that can be seen by the human eye and stored in its neurons. Let X be a black dot with a white background. The human brain stores the X in a neuron.  Many Xs are gathered in order to structure an image and a label is assigned. Our ANN model works in the same way. The central mechanism splits the input image into the smallest storable values i.e. pixels to be stored into the neurons, in the same way as a human brain divides an image into Xs. 

\begin{figure}[!t]
    \centering

        \includegraphics[width = 3.25in]{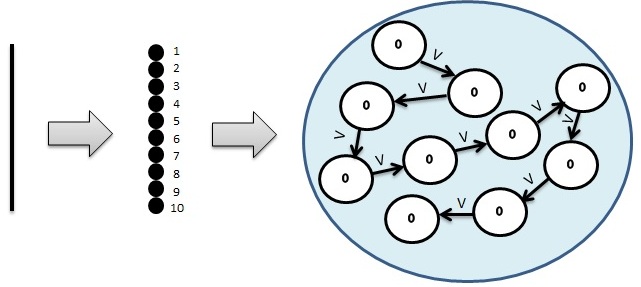}
     
     \caption{Line segment of 10 pixels transformation to subnet of 10 neurons}
        \label{LF}
\end{figure}

Let us proceed further by getting assistance from the example of a line. We take a straight vertical line segment of 10 pixels, as illustrated in Figure \ref{LF}. Our model stores these pixels along with the location of each pixel from the connected pixels. It is important to record the place of each pixel to structure the image. We have a subnet of 10 neurons to store the pixel values: one pixel per neuron. The label of connection between the neurons defines the vertical direction V of a pixel from the connected pixels, as depicted in Figure \ref{LF}. We have 0 and 1 values for black and white, respectively.

\begin{figure}[!t]
    \centering

        \includegraphics[width = 3.25in]{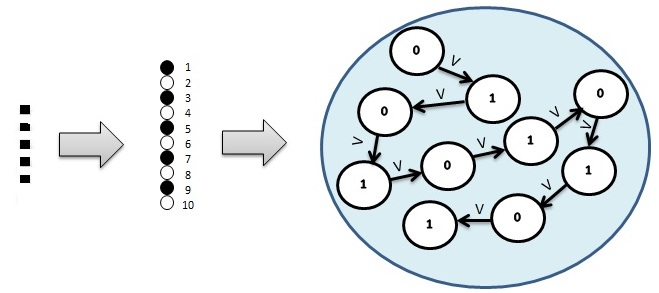}
     
     \caption{Dotted line segment of 10 pixels transformation to subnet of 10 neurons}
        \label{DLF}
\end{figure}

A group of pixels structuring an image can belong to more than one color. We take a black and white dotted-line segment to describe how our model stores the image having different colors. We have 0 and 1 values for black and white pixels, respectively, as depicted in Figure \ref{DLF}. Our model has a subnet of total 10 neurons (5 black and 5 white) to store the pixel values.

\begin{figure}[!t]
    \centering

        \includegraphics[width = 3.25in]{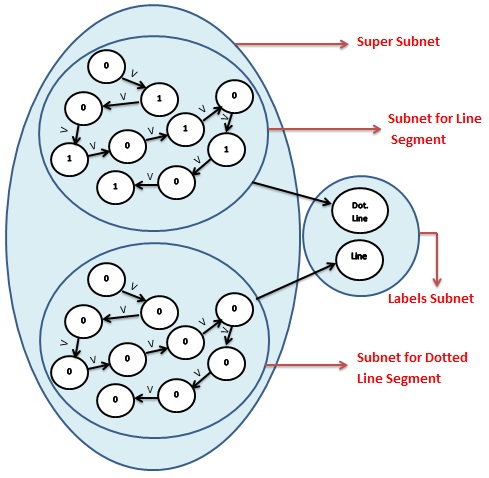}
     
     \caption{A super subnet contain the subnets of line and dotted line that are connected to the label neurons}
        \label{F10}
\end{figure}

We mentioned earlier that a subnet can have more than one related subnets, called super subnet. We can have many subnets for holding the shapes in our ANN, so we create a super subnet to hold all subnets defining the shapes. We have a super subnet holding two subnets of shapes: one for line and the other for dotted line, as illustrated in Figure \ref{F10}. Humans have given a particular name to every object in this universe, so the image stored in our ANN also has a label. In our model, the image subnet is connected to the neurons containing the labels. Subnets of line and dotted line are connected to the label neurons.

\subsection{Feature Extraction}

For example, we have an image which includes many shapes like line, dotted line, circle, and various other shapes.  Our ANN model will filter out all the shapes from the image and make a super subnet for an image which will have a subnet against every single shape in the image to store. In other words, every shape that we filter from the image is a feature so each subnet in the super subnet of the image is actually a feature. Moreover, every feature has a label, like subnets of line and dotted line, are two features and connected to the label neurons of "line" and "dotted line", respectively, shown in Figure \ref{F10}.

\subsection{Testing}
\begin{figure}[!t]
    \centering

        \includegraphics[width = 3.25in]{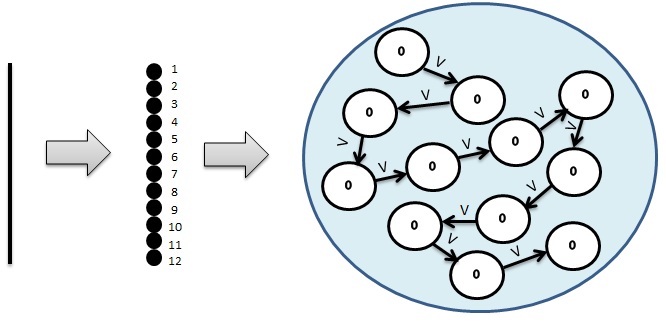}
     
     \caption{Line segment of 12 pixels transformation to test subnet of 12 neurons}
        \label{F12}
\end{figure}

In image recognition, we have to select the appropriate subnet(s) that best describes the test image. We take an image of a vertical straight-line segment of 12 pixels as a test image. Our ANN model splits and stores the test image in the same way as in the training image, and then compares with the subnets in ANN. So, 12-pixel input image of the line segment is stored into a subnet of 12 neurons of black color called test subnet, given in Figure \ref{F12}. The test subnet is compared with both subnets: line subnet and dotted line subnet, as given in Figure \ref{F10}. \textit{Comparison is made on the basis of the number of same color neurons and the label of connections between them.} The line subnet and dotted line subnet have 10 and 5 black color neurons, respectively, and both have the same label of connections between their neurons, that is "vertical" (V). Our test subnet has 12 black neurons that are near to the line subnet having 10 black color neurons, so the line subnet best defines the test image.

If we add the flexibility of the line direction such that the line can be in any direction, then we just need to change the label of connection between the neurons. The label of the connection between the first and second neuron is "any direction" (AD). The term "any direction" means the direction may be vertical, horizontal and so on. It is important to note that rest of the neurons in a given subnet will have the same type of direction and we call it "Inherit direction" (IA). For example, if the direction between neuron X and Y is vertical, the direction of all the neuron in that particular subnet will also be vertical.  The subnets containing the basic shapes which make the base of images, they can be considered as a prior knowledge, for example, a common subnet for a line in any direction. So, in a real-time image processing, all super subnets containing the basic shapes will act as the prior knowledge.

\begin{figure}[!t]
    \centering

        \includegraphics[width = 2.25in]{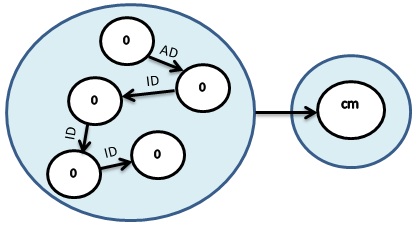}
     
     \caption{The structure of the subnet contains prior knowledge of one cm}
        \label{F11}
\end{figure}

If we need to compare an image of circle drawn on the page with an image of London Eye, which is actually a large circular shape, even then the testing process will remain the same. As both images are analyzed based on the pixels and, probably have the same size, and the most important point is that we are not measuring the length or width of the objects. However, we also need to measure the objects. We take a very simple, yet powerful real-time example to understand, how our model measures the objects. If you are shown a tree and asked its height, your answer may be \textit{8 feet. How long a foot is?} You may answer \textit{12 inches.} \textit{How long an inch is?} You will refer to some other parameter to define the inch and you may answer \textit{2.54 cm} and so on. In this world, we measure an object by referring to some other object. International standards have been defined of the scale units to calculate the things like centimeters, inches, and feet etc. Technically, a scale is a well-defined collection of dots or pixels. We need such type of information as a prior knowledge to accurately analyze the images. For example, if we store the prior knowledge of one cm and assume that it takes 5 pixels in a sequence in any direction. The structure of the subnet for one cm is depicted in Figure \ref{F11}.

\begin{figure}[!t]
    \centering

        \includegraphics[width = 3.5in]{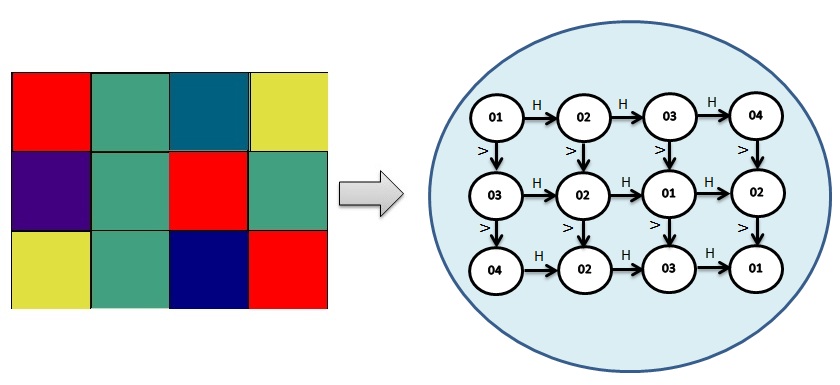}
     
     \caption{3*4 colorful image transformation to subnet}
        \label{F13}
\end{figure}

The working of our ANN model for colorful images will remain the same as for black and white images. We can replace the black and white contrast with any X and Y color contrast. Our model can have two approaches for colorful image recognition. First, convert the image into grayscale and recognize the shapes and then colors. A second approach would be to directly process the colorful image. Figure \ref{F13} illustrates an image having four colors: red, green, blue, and yellow and we used values, 01, 02, 03, and 04, respectively in the neuron for transformation to our ANN. We have a total of 12 neurons in the subnet. The first neuron contains red color code i.e. 01 and connected with 2 neurons: first is in the horizontal direction (green) and second in the vertical direction (blue). We can also use x, y and z axis or any other mechanism to identify the directions.

\subsection{Handwritten Digits Recognition Example}

\begin{figure}[!t]
    \centering

        \includegraphics[width = 2.5in]{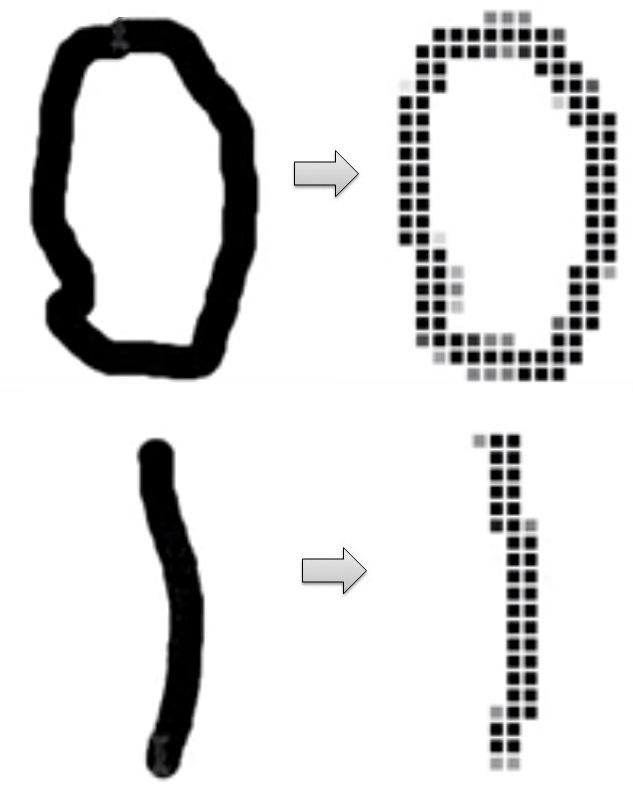}
     
     \caption{Training images representing the value 0 and 1}
        \label{F15}
\end{figure}

\begin{figure}[!t]
    \centering

        \includegraphics[width = 2in]{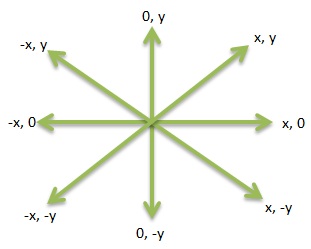}
     
     \caption{8 possible directions of a neuron from interconnected neurons}
        \label{F16}
\end{figure}

We train our model with 2 grayscale images representing the handwritten values 0 and 1: each value per image. The training images contain 118 and 43 black pixels representing the value 0 and 1, respectively, as shown in Figure \ref{F15}.  We have subnet 1 and 2 containing 118 and 43 neurons, respectively, to hold these two images. Subnet 1 and 2 are also connected to the label neurons 0 and 1, respectively, as the line and dotted line subnet are connected to label neurons in the previous example, as illustrated in Figure \ref{F10}. For the grayscale images, the pixel value represents the brightness of the pixel. The most common pixel format is the byte image, where this number is stored as an 8-bit integer giving a range of possible values from 0 to 255. For this example, 0 is taken to be black, and 1 is taken to be white. Our training images are 2D images, so the place of a neuron with the interconnected neurons is defined in x and y axis. We can have 8 possible directions of a neuron from interconnected neurons: (x, 0), (0, y), (-x, 0), (0, -y), (x, y), (-x, -y), (x, -y), and (-x, y), as depicted in Figure \ref{F16}.

\begin{figure}[!t]
    \centering

        \includegraphics[width = 2in]{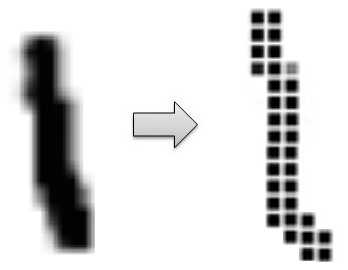}
     
     \caption{Test image representing the value 1}
        \label{F17}
\end{figure}

We take an image representing the value 1 as a test image.  We have to select the appropriate subnet from subnet 1 and 2 that best describes the test image. Our ANN model splits and stores the test image in the same way as in the training image, and then compares with the subnets in ANN. The test image contains 33 black pixels, shown in Figure \ref{F17}.  The comparison is made on the basis of the number of same color neurons and the label of connections between them. The subnet 1 and 2 has 118 and 43 black color neurons, respectively.  Our test subnet has 33 black neurons that are near to the subnet 2 (representing value 1, which is the accurate value) having 43 black neurons, so the subnet 2 best defines the test image.

We can take initiative from any pixel in the image to transform the image into the subnet, however, the same pattern should be followed for both, training and testing images. For this example, we start from the leftmost pixel in the first row and then proceed in rows, for both train and test images. The subnet 1 has 118 neurons which are more than the double of the neurons in subnet 2; subsequently, subnet 1 has more connections as compared to the subnet 2. The connection labels between the neurons of subnet 1 and subnet 2 are in the direction of (x, 0), (0, –y) and (x, -y) axis. However, the subnet 1 has the circular shape of pixels representing the 0, so it also contains the (x, y) and (–x, -y) connection labels. The connection labels between the neurons of our test subnet are in (x, 0), (0, –y) and (x,-y) directions. Test image doesn't have any circular shape of pixels so don't contain the (x, y) and (–x, -y) labels and near to the subnet 2. So the subnet 2 best defines the test image.

\subsection{Real-Time 3D Image Processing}

We are living in a 3D world however our eyes can see only two dimensions.  The sense of depth perception or determining the distance is a trick that our brains have learned. The brain learns this by putting together two (from two eyes) 2D images in such a way as to extrapolate the distance, called stereoscopic vision. This is one of the very basic learning that a newborn baby learns immediately after the birth however this is the key challenge to cramming this knowledge into machines. On behalf of this learning experience, the humans are efficient enough to easily recognize the 2D images captured on papers.

The procedure for real-time 3D image processing will remain the same however we need the location of pixels that are structuring an image in the 3D world. We can apply multiple techniques in order to find the location of the pixels of 3D images. Mapping 2D images onto the 3D image is an active research area. We can put together multiple 2D images to extrapolate the location of the pixels, like stereoscopic vision. Alternatively, we can use sensors (for example, Infrared (IR) sensors) along with the camera to capture the real-time 3D scenery, in order to define the exact location of each pixel.

\section{Literature Review}

Although the concept of ANN is not new however the implementation has only been possible over the last couple of years. This paper \cite{krizhevsky2012imagenet} is widely regarded as one of the most influential publications in the field of deep neural networks. The authors proposed network architecture called Alexnet, made up of 5 layers, max-pooling layers, dropout layers, and 3 fully connected layers. The network used for classification with 1000 possible categories. In \cite{zeiler2014visualizing}, the author proposed a slightly modified AlexNet model and a way of visualizing feature maps. They also talk about the limited knowledge that researchers had on inner mechanisms of these models. In another influential paper \cite{simonyan2014very}, the authors gave the concept of \textit{keep deep and keep simple}. The authors proposed a 19 layer CNN that strictly used 3x3 filters with stride and pad of 1, along with 2x2 maxpooling layers with stride 2. In \cite{szegedy2015going}, the authors also promoted the idea of simplicity and depth and proposed a network GoogLeNet contains 22 layers and was the winner of ILSVRC 2014 with a top 5 error rate of 6.7\%. ResNet is a more deep network with 152 layer network architecture that Microsoft Research came up with in late 2015 \cite{he2016deep}. 
In \cite{karpathy2014deep}, the authors combined the CNNs with RNNs models to create a useful application that in a way combines the fields of Computer Vision and Natural Language Processing. 

Faster R-CNN has become the standard for object detection programs today. The purpose of R-CNNs is to overcome the issue of object detection. It draws the bounding boxes over all of the objects in a certain image \cite{girshick2014rich}. Fast R-CNN \cite{girshick2015fast} and faster R-CNN \cite{ren2015faster} proposed to make the model faster and better suited for modern object detection tasks. In fast R-CNN, improvements are made to the original model and faster R-CNN works to combat the somewhat complex training pipeline that both R-CNN and Fast R-CNN exhibited. The region proposals by R-CNN are then “warped” into an image size and fed into a trained deep neural network that extracts a feature vector for each region.  A new way of transformation called spatial transformation is proposed in \cite{jaderberg2015spatial}, the authors implemented the simple idea of making affine transformations to the input image in order to help models become more invariant to translation, scale, and rotation.

Although deep neural networks have been used for image recognition for a long time, but in order to mimic the biological neural networks and enhance the performance we need to rethink to understand the deep learning methods \cite {zhang2016understanding} new paths are needed to artificial intelligence like in \cite{bottou2014machine}, the author replaced the rule-based manipulation of symbolic expressions by operations on large vector. In \cite{szegedy2013intriguing}, the authors applied an imperceptible non-random perturbation to an image and considered a state-of-the-art deep neural network which arbitrarily changes the network’s prediction. These adversarial examples definitely surprised the researchers and quickly became a topic of interest. 

\section{Discussion}

\begin{figure}[!t]
    \centering

        \includegraphics[width = 3.5in]{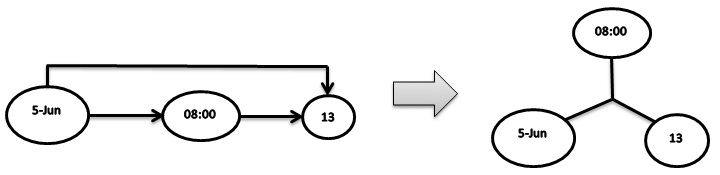}
     
     \caption{Connection is refined and converted into single connection connecting three neurons}
        \label{F14}
\end{figure}

This is the first work in ANN domain which talks about a non-layered architecture with the central mechanism, so we have tried to make it as simple as possible by removing complexities. When we talk about the connections between the neurons, we limit it between two neurons. In fact, a single connection can connect to multiple neurons. Figure 7 contains an image which that shows the insulin pump dataset. We redefined the connections and combined them to form a single connection interconnecting the three neurons, as shown in Figure \ref{F14}.

The neurons and subnets in an ANN is a storage mechanism, whereas the central mechanism is the main processing unit which has control over how deeper to process the information. For example, the central mechanism can process the test record by finding 3 nearest values of input record (in the given examples we are limited to find one nearest value) and calculate their average. The in-depth processing is also capable to process bias more deeply. All neurons are interconnected and have direct access to all other neurons in the ANN through the central mechanism. In in-depth processing, the central mechanism may pass the input data directly from one neuron to any other neuron in the whole ANN. Backpropagation is also performed by the central mechanism.

In this paper, we explained our proposed model as simply as much possible to make it more accessible to everyone. We need to involve mathematics, statistics and some other fields. For example, we can use mathematical schemes during the testing phase where the central mechanism takes input and forward it to the appropriate subnets. Moreover, parallel processing can be performed in ANN to find the appropriate subnets to process the data. Mathematics can be very useful in many other ways for the simplification of our model. For example, logarithms can be used for representing very large amounts in a brief form.

\section{Conclusion}

We proposed a simple, yet powerful ANN model truly inspired by the human neural network. Our ANN architecture contains a mesh of subnets controlled by a central mechanism. A subnet is a group of neurons, where a neuron acts as memory cells to hold the dataset values. The central mechanism is responsible for making decisions, controlling and monitoring all the capabilities of the ANN. To the best of our knowledge, this is the first work that proposes an ANN model based on a combination of behavioral observation of human intelligence and scientific research on human neurons. We proposed a novel approach for textual data training and testing process by using examples. Moreover, we introduced a new way of storing and processing the images in our ANN model so that it can be best used to compare with the test image and useful in image recognition. In the future, we intend to make a significant improvement to the feature extraction capabilities of the ANN by focusing on the ways to enhance the quality of training data rather than focusing on its quantity.

\appendices
\section{Insulin Pump Dataset}

\begin{table}[!t]
\renewcommand{\arraystretch}{1.3}
\caption{Insulin Pump Dataset}
\centering
\begin{tabular}{|c|c|c|c|}
\hline
\bfseries Sr.\# & \bfseries Date & \bfseries Time &  \bfseries Insulin Amount \\ 
\hline \hline
1                                     &        1-June       &      08:00         &               13          \\ \hline
2                                     &         1-June	      &     18:00	          &         32                \\ \hline
3                                     &          2-June     &         10:00       &           12              \\ \hline
4                                     &          2-June     &          17:00      &             30            \\ \hline
5                                     &           3-June    &         07:00       &               14          \\ \hline
6                                     &          3-June     &         16:00       &                 32        \\ \hline
7                                     &          4-June     &        10:00        &                   12      \\ \hline
8                                     &          4-June     &          17:00      &                     34    \\ \hline
9                                     &             5-June     &         08:00      &                     12    \\ \hline
10                                    &             5-June     &        18:00        &                      32   \\ \hline
11                                    &             6-June     &      11:00          &                        13 \\ \hline
12                                    &     6-June     &           17:00     & 30                         \\ \hline
13                                    &     7-June     &           08:00     & 11                        \\ \hline
14                                    &     7-June     &         19:00       &   32                      \\ \hline
15                                    &     8-June     &        07:00        &     12                    \\ \hline
16                                    &     8-June     &      18:00          &       34                  \\ \hline
17                                    &     9-June     &          07:00      &         11                \\ \hline
18                                    &     9-June     &          20:00      &           30              \\ \hline
19                                    &     10-June     &        06:00        &            13             \\ \hline
20                                    &     10-June     &        17:00        &              35           \\ \hline
21                                    &     11-June     &       10:00         &                14         \\ \hline
22                                    &     11-June     &      18:00          &                  32       \\ \hline
23                                    &     12-June     &  09:00                  &                12         \\ \hline
24                                    &     12-June     &       19:00        &  16                       \\ \hline
25                                    &     13-June     &  09:00             &    12                     \\ \hline
26                                    &     13-June     &            16:00   &      32                   \\ \hline
27                                    &     14-June     &          08:00     &        13                 \\ \hline
28                                    &     14-June     &        21:00       &          30               \\ \hline
29                                    &     15-June     &       10:00        &            12             \\ \hline
30                                    &     15-June     &      20:00         &              34           \\ \hline
\end{tabular}
\end{table}

\bibliographystyle{IEEEtran}
\bibliography{ref}

\begin{IEEEbiography}[{\includegraphics[width=1in,height=1.25in,clip,keepaspectratio]{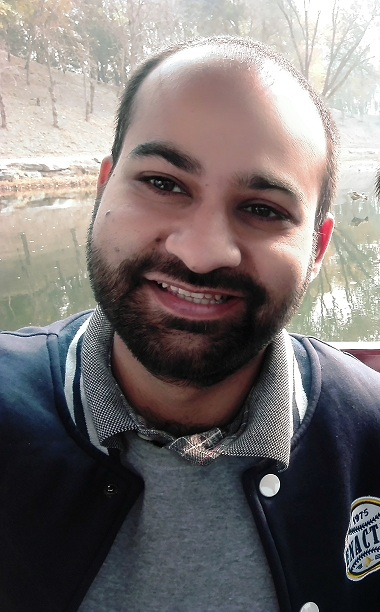}}]{Usman Ahmad}
received his MS (Computer Science). He is currently pursuing the Ph.D. studies in the School of Computer Science and Technology, Beijing Institute of Technology, Beijing, China. His research focuses on machine learning, deep learning, artificial neural networks, information security and internet of things.
\end{IEEEbiography}

\begin{IEEEbiography}[{\includegraphics[width=1in,height=1.25in,clip,keepaspectratio]{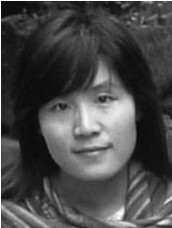}}]{Professor Song Hong}
's main research areas are computer graphics image processing, medical image processing, and pattern recognition. She has published over 30 academic papers and presided several research projects including the National Natural Science Foundation of China and the National 863 Program.
\end{IEEEbiography}

\begin{IEEEbiography}[{\includegraphics[width=1in,height=1.25in,clip,keepaspectratio]{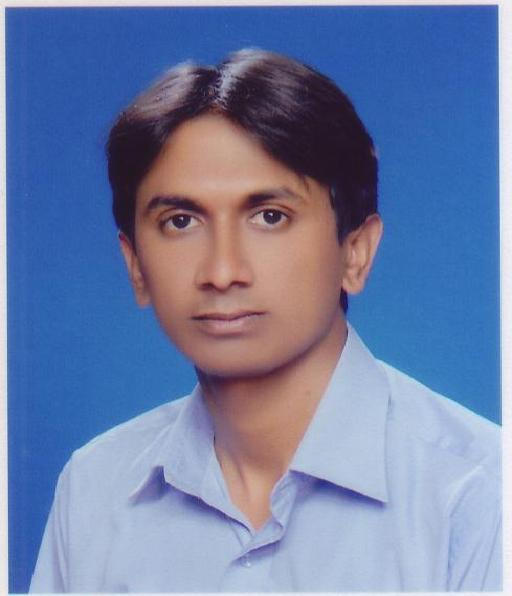}}]{Awais Bilal} received his MS (Information Security) degree from School of Electrical Engineering and Computer Science (SEECS), National University of Sciences and Technology (NUST), Pakistan. His research interests include information security, digital forensics and cloud computing security.
\end{IEEEbiography}

\begin{IEEEbiography}[{\includegraphics[width=1in,height=1.25in,clip,keepaspectratio]{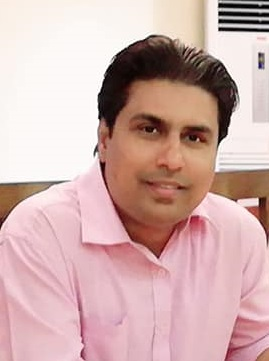}}]{Shahid Mahmood} is pursuing his Ph.D. studies in Automotive Cybersecurity at Centre for Future Transport and Cities, Coventry University, UK. He received his master's degree in Software Development (with distinction) from Coventry University, UK. His areas of research interest mainly focus on cyber security with a particular focus on automotive and mobile security. 

\end{IEEEbiography}

\begin{IEEEbiography}[{\includegraphics[width=1in,height=1.25in,clip,keepaspectratio]{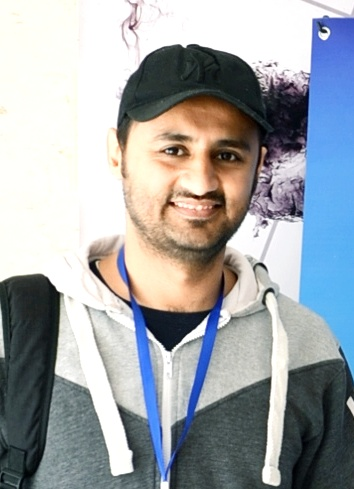}}]{Asad Ullah}
is pursuing Ph.D. (Information and Communication Engineering) School of Information and Electronics, Beijing Institute of Technology and on Study leave at Riphah International University, Faisalabad, Pakistan where he was serving as Assistant Professor. His area's of research are Digital Image, Processing, Computer Vision and Pattern Recognition etc and having several publications in these areas. 
\end{IEEEbiography}

\begin{IEEEbiography}[{\includegraphics[width=1in,height=1.25in,clip,keepaspectratio]{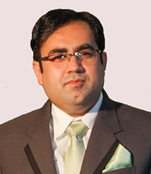}}]{Uzair Saeed}
received M.sc. degree in Computing Systems from Department of Computing, Nottingham Trent University, Nottingham, UK. He is currently pursuing
the Ph.D. degree in Computer Science and Technology in the School of Computer Science and Technology, Beijing Institute of Technology, Beijing, China. His main research areas are human computer interaction, augmented reality, deep computer vision and facial expression.
\end{IEEEbiography}

\end{document}